\documentclass[lettersize,journal]{IEEEtran}
\usepackage{amsmath,amsfonts}
\usepackage{algorithmic}
\usepackage{algorithm}
\usepackage{array}
\usepackage[caption=false,font=normalsize,labelfont=sf,textfont=sf]{subfig}
\usepackage{textcomp}
\usepackage{stfloats}
\usepackage{url}
\usepackage{verbatim}
\usepackage{graphicx}
\usepackage{accessibility}
\usepackage[numbers]{natbib}
\usepackage{accessibility}

\usepackage{xcolor,colortbl}
\definecolor{myorange}{HTML}{FF8000}
\usepackage{multirow}
\usepackage{booktabs}
\usepackage{graphicx} % for \rotatebox
\usepackage{textgreek}
\usepackage{pifont}
\usepackage{tcolorbox}
\usepackage{paralist}
\usepackage{soul}
\usepackage{orcidlink}
\usepackage[T1]{fontenc}
\usepackage[utf8]{inputenc}
\tcbuselibrary{breakable}
\newtcolorbox{example}[1][]{
  breakable,
  title=#1,
}
\usepackage{threeparttable}

\begin{document}

\title{ContextClaim: A Context-Driven Paradigm for Verifiable Claim Detection}

\author{Yufeng Li\,\orcidlink{0009-0008-8740-4994}, Rrubaa Panchendrarajan\,\orcidlink{0000-0002-1403-2236}, Arkaitz Zubiaga\,\orcidlink{0000-0003-4583-3623}%,~\IEEEmembership{Member,~IEEE,}
        % <-this % stops a space
\thanks{Yufeng Li, Rrubaa Panchendrarajan, and Arkaitz Zubiaga are with the School of Electronic Engineering and Computer Science, Queen Mary University of London, London E1 4NS, U.K.}

}



\maketitle

\begin{abstract}
Automated fact-checking pipelines typically begin with a filtering stage that decides which claims are worth verifying, given that the later evidence retrieval and verification components are expensive to apply at scale. A central task in this stage is verifiable claim detection, which asks whether a statement is in principle checkable against external evidence. Prior work on this task, as well as on the closely related notion of check-worthiness, conditions its decisions only on the claim sentence itself. We argue that this is restrictive, because deciding whether a statement is checkable often depends on identifying the entities and events it mentions, and on whether external information about them is actually available in the first place. Motivated by how downstream verification systems rely on retrieved evidence, we move retrieval upstream into the detection stage and introduce ContextClaim. Given an input claim, the approach identifies entity mentions, queries Wikipedia as a structured background source, and uses large language models to compress the retrieved material into short contextual summaries that are then passed to a classifier. Experiments are conducted on two domains and genres, namely the CheckThat! 2022 Twitter collection and the PoliClaim corpus of political debates, and cover both encoder and decoder only models under fine-tuning, zero-shot, and few-shot settings. The added context yields gains on verifiable claim detection in several configurations, although the size of the improvement varies with the dataset, the backbone model, and the training setup. We further find that the same retrieved summaries are useful beyond detection. Feeding them into a downstream verification model on FEVER improves verification F1. Component level analyses, human annotation, and error inspection further clarify the conditions under which retrieved context helps, and where it does not.
\end{abstract}


\section{Introduction}\label{sec:intro}
Misinformation spreads rapidly across online platforms, shaping public opinion, distorting democratic processes, and amplifying social harm \citep{zubiaga2018detection,oladokun2024misinformation}. As a result, automated fact-checking (AFC) has become increasingly important for supporting human fact-checkers and improving the scalability of verification workflows \citep{thorne2018FactExtraction,zeng2021Automatedfact}. AFC is commonly described as a multi-stage pipeline involving (i) claim detection, (ii) matching against previously fact-checked claims, (iii) evidence retrieval, and (iv) claim verification. Early-stage filtering in the claim detection step by getting rid of claims that are not verifiable is especially important because it helps remove content that does not need further processing, thereby reducing the burden on downstream components \citep{shaar2020known}.

Claim detection, the first stage of this pipeline, aims to identify statements that warrant further fact-checking. Early research defined this task primarily through subjective prioritization criteria. Check-worthiness detection, the most widely studied formulation, prioritizes claims based on perceived public importance or interest \citep{micallef2022true, das2023state}, while related efforts target claims that may cause social harm or attract undue attention \citep{shaar2021findings, nakov2022overview}. However, these criteria are inherently subjective, as what counts as important, harmful, or attention-worthy varies across institutions, regions, and cultural contexts, and shifts over time as events evolve and public priorities change, limiting the generalizability of models trained under such criteria. To address this, recent work has proposed \emph{verifiable claim detection} \citep{konstantinovskiy2021AutomatedFactchecking}, which shifts the focus to a more objective criterion by asking whether a claim is \emph{checkable}, that is, whether it expresses a factual assertion that can, in principle, be assessed against external evidence \citep{alam2021fighting, ni2024afacta}. This framing reduces reliance on subjective judgment and offers a more generalizable foundation for claim detection.

However, across both check-worthy and verifiable claim detection, existing approaches share a common limitation in that they rely solely on the claim text itself, whether through linguistic heuristics and surface-level features \citep{dhar2019hybrid, favano2019theearthisflat, williams2020accenture, wuhrl2024makes} or, more recently, large language models (LLMs) using in-context learning and fine-tuning \citep{sawinski2023openfact, li2024factfinders, ni2024afacta}. For verifiable claim detection specifically, this leaves a notable gap. If verifiability is defined by whether a claim \emph{can be checked against evidence} \citep{gupta2021lesa, ni2024afacta}, then retrieved information about the entities and events mentioned in the claim could directly support this judgment. Inspired by the established role of evidence retrieval in later-stage claim verification \citep{thorne2018FactExtraction}, where external information is used to assess a claim's truthfulness, we explore whether a similar retrieval step can be applied earlier in the pipeline, not to verify claims, but to help determine whether they are verifiable in the first place. This leads us to reframe the task so that a claim is considered verifiable if it makes specific factual statements that can be checked against evidence, and retrieved contextual information can serve as a basis for making this judgment.

This idea is particularly relevant when the claim text alone does not reveal enough about the entities it mentions. For example, the statement ``Phineas and Ferb would have made the vaccine by now'' contains named entities, but background knowledge is needed to recognize that they are fictional characters, which makes the claim unverifiable as a real-world factual assertion. By contrast, ``I'm a nurse and Lindsey Graham got a vaccine before me'' also contains named entities, but retrieving contextual information confirms that it refers to an identifiable public figure and a real-world event, making the claim more plausibly verifiable. In both cases, the claim text itself appears specific, yet the verifiability judgment depends on external knowledge about the entities involved.

Motivated by this observation, we propose \textbf{Context-Driven Claim Detection (ContextClaim)}, a paradigm that augments verifiable claim detection with retrieved background context. ContextClaim collects contextual information from Wikipedia based on entity mentions in the input claim and uses LLMs to summarize that information for downstream classification. By bringing retrieval forward to the detection stage, the retrieved context directly supports the verifiability judgment. We further demonstrate that the same context transfers to downstream verification on FEVER \cite{thorne2018FactExtraction}, supporting a cross-stage view of retrieval reuse in automated fact-checking pipelines. A central question guiding our analysis is whether and under what conditions such context augmentation improves verifiable claim detection, as its effectiveness may vary depending on factors such as domain characteristics, model architecture, and the quality of retrieved information.

\begin{figure}[t]
    \centering
    \includegraphics[width=0.9\linewidth]{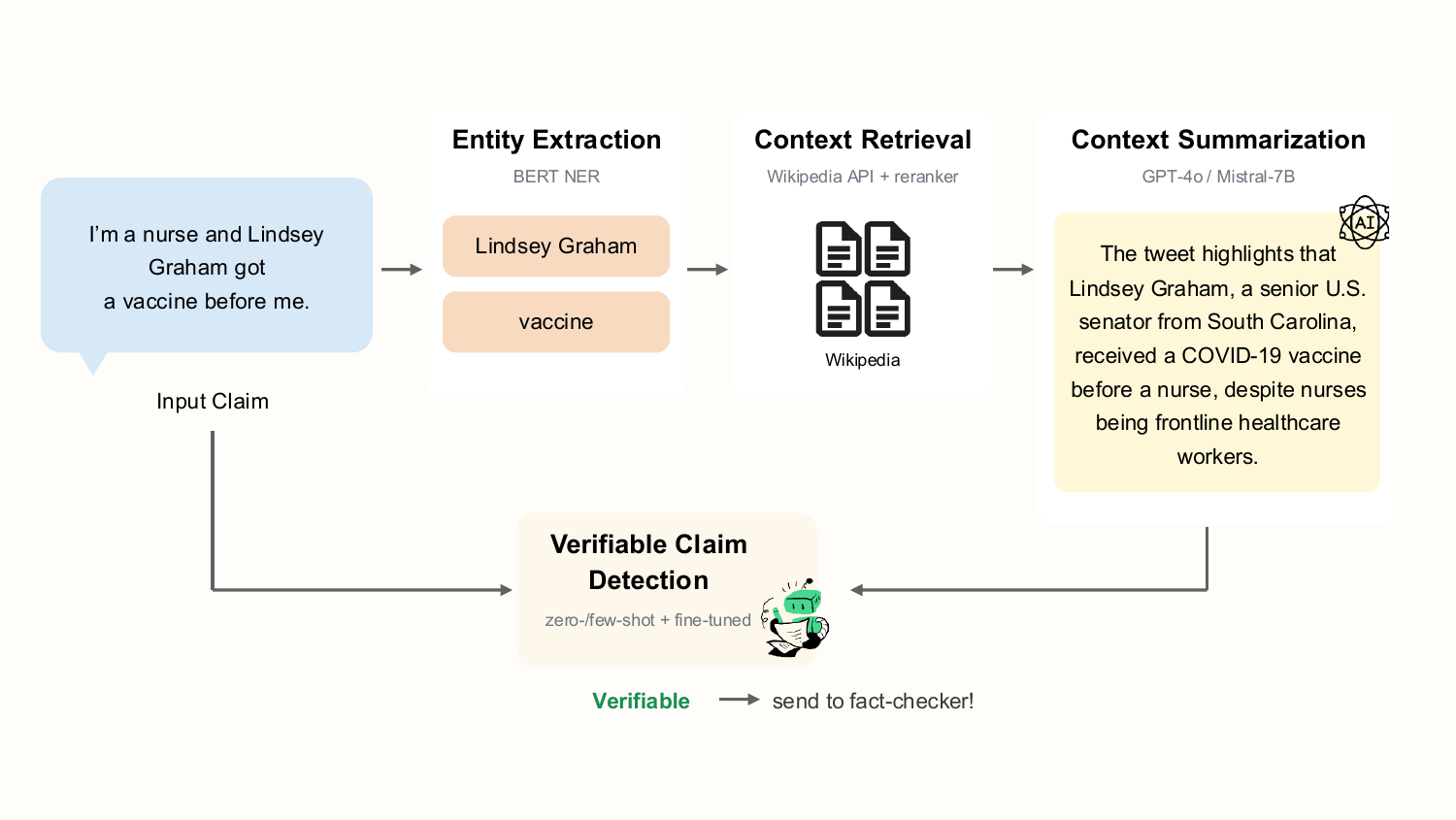}
    \caption{Example of ContextClaim system processing a vaccine claim about Lindsey Graham.}
    \label{fig:afc-illustration}
\end{figure}

Our paradigm comprises four components: entity extraction, context retrieval, context summarization, and verifiable claim detection. Figure~\ref{fig:afc-illustration} illustrates the full pipeline with a concrete example.

To evaluate ContextClaim, we conduct experiments on two datasets with different topical and genre characteristics: the English CheckThat! 2022 COVID-19 dataset \citep{nakov2022overview} (hereafter CT22) and the PoliClaim dataset. Importantly, our study focuses on datasets annotated for \emph{verifiability}. This setup allows us to examine verifiable claim detection across both COVID-19-related social media content and political discourse. We evaluate ContextClaim across encoder-only and decoder-only models under fine-tuning, zero-shot, and few-shot settings.

Our contributions are summarized as follows:

\begin{itemize}
\item We propose ContextClaim, a paradigm for augmenting verifiable claim detection with retrieved background context and LLM-based summarization at an early stage of the AFC pipeline.
\item We provide a systematic evaluation of context augmentation for verifiable claim detection across two datasets, multiple model architectures, and three learning settings: fine-tuning, zero-shot, and few-shot.
\item We identify the conditions under which context augmentation improves verifiable claim detection, and characterize the factors that influence its effectiveness, including domain, model type, and learning setting.
\item We complement the main experiments with component analysis, human evaluation, and error analysis, providing a detailed understanding of the role that retrieved context plays in the detection process.
\item We evaluate the cross-task utility of ContextClaim by applying its retrieved context to claim verification on FEVER, showing that context designed for detection also benefits downstream verification.
\end{itemize}

Our code, prompts, and configurations are released to support reproducibility.\footnote{Repository: \url{https://github.com/isyufeng/contextclaim}} Our results indicate that enriching claims with retrieved context improves verifiable claim detection in the majority of configurations, but that its effectiveness depends on how stably a given model can integrate additional information into its decision process. These findings highlight the importance of aligning retrieval and summarization strategies with the downstream model's capacity for context integration, a consideration that has received limited attention in prior work on claim detection.

\section{Related Work}\label{sec:related_work}

\subsection{Claim Detection}

Claim detection research has traditionally focused on claim check-worthiness estimation \citep{kartal2023ReThinkYou}, which ranks or classifies claims based on criteria such as public importance, social harm, or public interest \citep{panchendrarajan2024claim, micallef2022true, das2023state, shaar2021findings, nakov2022overview}. However, these criteria are often subjective and may vary across domains, audiences, and time. Recent work therefore shifts towards verifiable claim detection, where the goal is to determine whether a statement expresses a factual assertion that can in principle be checked against external evidence \citep{konstantinovskiy2021AutomatedFactchecking, gupta2021lesa}. Existing systems typically rely on the claim text alone, using linguistic features, pretrained encoders, or LLM-based prompting and fine-tuning \citep{hassan2017ClaimBusterfirstever, jaradat2018ClaimRankDetecting, dhar2019hybrid, williams2020accenture, agresti2022polimi, li2024factfinders}.  AFaCTA~\citep{ni2024afacta} is particularly relevant, as it uses multi-step LLM reasoning to assist verifiable claim annotation and introduces the PoliClaim dataset. Nevertheless, it still relies on the LLM's parametric knowledge rather than externally retrieved information. RAVE~\cite{li2026rave} is closely related, as it augments verifiable claim detection with entity-based web retrieval and relevance-credibility scoring. ContextClaim differs by using Wikipedia as a structured knowledge source, transforming retrieved information into LLM-generated contextual summaries, and systematically evaluating how such context affects different model architectures, learning settings, and downstream verification utility.

\subsection{Evidence Retrieval and Context-Augmented Fact Checking}
Evidence retrieval is commonly used in later stages of automated fact-checking, where systems retrieve documents or rationales to support veracity prediction~\citep{guo2022SurveyAutomated}. FEVER~\citep{thorne2018FactExtraction} established Wikipedia-based evidence retrieval as a standard setting, and subsequent work has combined sparse retrieval methods such as TF-IDF \citep{ramos2003using} and BM25 \citep{robertson2009probabilistic} with neural retrievers, rerankers, and generative evidence retrieval models~\citep{hanselowski2018ukp, soleimani2020bert, chen2022gere, zheng2024evidence}. More recent retrieval-augmented generation (RAG) approaches integrate external context into LLM-based fact verification without full model retraining~\citep{yue2024evidence}. For example, HF-RAG combines labeled examples and unlabeled Wikipedia evidence from multiple rankers through hierarchical fusion, showing that multi-source and multi-ranker retrieval can improve fact verification and out-of-domain generalization~\cite{santra2025hf}. These findings support the broader view that retrieved context can provide task-relevant grounding beyond the input claim itself.

At the same time, recent work shows that retrieval quality and context utilization remain major bottlenecks. \citep{hagstrom2025reality} introduces DRUID, a real-world claim verification dataset with automatically retrieved and manually annotated contexts, and show that synthetic context-utilization benchmarks often fail to reflect the complexity, insufficiency, and noise of real retrieved evidence. Complementarily, \citep{wu2025fact} finds that modern retrievers and rerankers may rely heavily on semantic similarity rather than factual reasoning, making retrieval vulnerable to factual distractors and paraphrasing. These studies suggest that retrieved context is useful but not automatically reliable. ContextClaim differs from prior retrieval-based fact-checking systems by moving retrieval from the verification stage to the claim detection stage. Rather than retrieving evidence to support or refute an already selected claim, it retrieves background information about entities and events to help determine whether the claim is verifiable in the first place.

\section{Methodology}\label{sec:methodology}
We introduce ContextClaim, a context-driven paradigm designed to enhance claim detection by leveraging contextual information from Wikipedia. The paradigm operates through a sequence of components: (1) entity extraction: given an input claim $x_i$, the paradigm first identifies a set of named entities $E_i = \{e_{1}, e_{2}, ..., e_{m}\}$, then (2) context retrieval: for each extracted entity, the system retrieves relevant information from Wikipedia, selecting the most pertinent extracts $a_i$. These extracts are then reranked and aggregated to construct a comprehensive knowledge base $K_i$. (3) Context Summarization: since the raw knowledge base $K_i$ often contains redundant or irrelevant content, we summarize it conditioned on the original claim to produce a context summary, thereby suppressing noise and distilling the most claim-relevant information as additional input for claim detection. (4) Verifiable claim detection: finally, both the original claim $x_i$ and the context summary $c_i$ are input into a pre-trained language model, which classifies whether the claim is verifiable or not. Because retrieval and summarization are separated from the detector, the resulting context can be generated once and reused across different detection models and later pipeline stages, as we demonstrate with downstream verification.

% While ContextClaim is designed for the detection stage, the retrieved context can also be reused in later pipeline stages; as we later demonstrate this empirically on downstream verification.

\subsection{Entity Extraction}
Entities in a piece of text often carry the most important information. By extracting these entities, we can convert unstructured input into a more structured form, facilitating the subsequent context retrieval. We use a BERT-based named entity recognition (NER) model \citep{DBLP:journals/corr/abs-1810-04805} fine-tuned to identify entities with four standard types: Person (PER), Location (LOC), Organization (ORG), and Miscellaneous (MISC).

Standard NER models trained on datasets such as CoNLL-2003 \citep{tjong-kim-sang-de-meulder-2003-introduction} define a fixed set of entity types (PER, LOC, ORG, MISC) that does not include domain-specific categories such as diseases. A frequency analysis of the CT22 training set revealed that terms like ``covid-19'' and ``coronavirus'' appear in a substantial proportion of claims but are consistently unrecognized by the NER model, despite being central to a dataset composed entirely of COVID-19-related content. To address this coverage gap, we extend the entity type set with an additional \textsc{Disease} type, defined by a small lexicon of five terms: \{\textit{covid-19}, \textit{coronavirus}, \textit{corona virus}, \textit{sars-cov-2}, \textit{covid}\}. Matched terms are added to the extracted entity set via exact string matching. This extension is minimal and specific to the domain characteristics of CT22. For the PoliClaim dataset, which covers a broader range of political topics, the standard NER types are sufficient and no additional customization is applied.

Formally, let \( X = \{x_1, x_2, \dots, x_n\} \) be a set of input claims. For each claim \( x_i \), the extractor identifies a set of entities \( E_i = \{e_{i,1}, e_{i,2}, \dots, e_{i,m_i}\} \), where each entity \( e_{i,j} \) is a tuple \( (w_{i,j}, t_{i,j}) \). Here, \( w_{i,j} \) is the entity token and \( t_{i,j} \in T \) is its type, with \( T = \{\text{PER}, \text{LOC}, \text{ORG}, \text{MISC}\} \) as the standard NER types. For the CT22 dataset, the type set is extended to \( T' = T \cup \{\text{DISEASE}\} \) to incorporate disease-related entities identified through lexicon matching \citep{zobel1995finding}.

\subsection{Context Retrieval}\label{sec:method_retrieval}
For each extracted entity \( e_{i,j} \), we query Wikipedia through its API\footnote{\href{https://www.mediawiki.org/wiki/API:Main_page}{MediaWiki API documentation}.} to retrieve a set of candidate article extracts:
\[
A_{i,j} = \{a_{i,j,1}, a_{i,j,2}, \dots, a_{i,j,p}\}
\]
where \( p \) denotes the number of candidates returned per query.
To identify the most relevant extract for each entity, we re-rank the candidates using a relevance score that combines two signals: the semantic similarity between the extract and the original claim \( x_i \), and the alignment between the Wikipedia article title and the entity mention \( w_{i,j} \). Both similarities are computed as cosine similarity over dense sentence representations \citep{wang2020minilm}. The final score is a weighted combination:
\[
\text{score}(a_{i,j,k}, x_i) = \alpha \cdot f(a_{i,j,k}, x_i) + \beta \cdot f(\text{title}_k, w_{i,j})
\]
where \(\alpha{=}0.8\) and \(\beta{=}0.2\) are fixed weights. The weighting reflects the assumption that semantic alignment between the extract and the full claim should serve as the primary relevance signal, while title matching acts as a secondary disambiguation cue. This formulation follows the common practice in hybrid retrieval of combining dense semantic and sparse lexical signals through weighted fusion \citep{luan2021sparse, bruch2023analysis}.

We select the highest-scoring extract as the representative context for each entity:
\[
a_{i,j}^* = \arg\max_{a_{i,j,k} \in A_{i,j}} \text{score}(a_{i,j,k}, x_i)
\]
Repeating this for all entities in \( x_i \) yields a set of top-ranked extracts:
\[
A_i^* = \{a_{i,1}^*, a_{i,2}^*, \dots, a_{i,m_i}^*\}
\]

We then apply a type-based filtering step to retain only entity--extract pairs that are likely to provide informative context. Entities of type PER, LOC, and ORG typically correspond to focused Wikipedia articles containing directly relevant factual information, such as a politician's career history or a country's governance structure, and are therefore retained unconditionally. In contrast, entities of type MISC (and DISEASE for CT22) often link to broader Wikipedia articles whose content may be less specific to the claim. For these entity types, we apply a relevance threshold \( \theta \) and retain only those extracts whose score meets this threshold. Let \( T_c = \{\text{PER, LOC, ORG}\} \) denote the set of core entity types. The filtered set is defined as:
\[
\hat{A}_i = \{a_{i,j}^* \in A_i^* \mid t_{i,j} \in T_c\}
\]
\[
\cup\; \{a_{i,j}^* \in A_i^* \mid t_{i,j} \notin T_c \wedge \text{score}(a_{i,j}^*, x_i) \geq \theta\}
\]
The remaining extracts form the contextual knowledge base for the claim:
\[
K_i = \bigcup_{a_{i,j}^* \in \hat{A}_i} a_{i,j}^*
\]
This knowledge base \( K_i \) is passed to the next stage for summarization.

\subsection{Context Summarization}\label{sec:context-summarization}
The contextual knowledge base \( K_i \) constructed in the previous step may contain lengthy or partially relevant extracts from multiple Wikipedia articles. To distill this information into a form suitable for downstream classification, we generate a context summary \( c_i \) for each input claim \( x_i \) via a generation function \( g \):
\[
c_i = g(x_i, K_i)
\]
The generation function \( g \) is implemented using an LLM prompted with task-specific instructions. The prompt directs the model to produce a factual summary of approximately 150 words that draws only on the content in \( K_i \) while referencing \( x_i \), with the aim of preserving faithfulness to the retrieved material and reducing hallucinated content. By conditioning on both the claim and the retrieved knowledge base, the summary is intended to foreground information that is contextually relevant to the verifiability judgment rather than providing a generic overview of the retrieved extracts.

To examine whether summarization quality affects downstream detection performance, we evaluate two language models for this component: GPT-4o \citep{gpt4o2025}, a state-of-the-art instruction-following model, and Mistral-7B-Instruct-v0.2 \citep{jiang2023mistral7b}, a lightweight open-source alternative. We refer to their respective outputs as ContextClaim-G4o (CC-G4o) and ContextClaim-M (CC-M). Comparing these two models allows us to assess the sensitivity of the overall pipeline to the capacity of the summarization component.

\subsection{Verifiable Claim Detection}
In the final step, verifiability assessment is formalized as a binary classification task. Given the input claim \( x_i \) and its context summary \( c_i \), a classifier \( h \) predicts:
\[
v_i = h(x_i, c_i)
\]
where \( v_i \in \{0, 1\} \) denotes verifiable (\( v_i = 1 \)) or non-verifiable (\( v_i = 0 \)). To understand how different model architectures and learning paradigms interact with context augmentation, we implement \( h \) using both encoder-only models (BERT, RoBERTa) and decoder-only models (Llama3, Mistral) across three learning settings: fine-tuning, zero-shot, and few-shot (Section~\ref{sec:model}). This design enables a systematic comparison not only of whether context augmentation improves verifiable claim detection, but also of the conditions under which different model types benefit from the additional contextual input.

\section{Experiment}

\subsection{Datasets}\label{sec:dataset}

\begin{table}[t]
\centering
\caption{Dataset Statistics. Veri = Verifiable, Non = Non-verifiable.}
\label{tab:dataset_stats}
\setlength{\tabcolsep}{8pt}
\begin{tabular}{l rrr rrr}
\toprule
\multirow{2}{*}{Split}
  & \multicolumn{3}{c}{CT22}
  & \multicolumn{3}{c}{PoliClaim$^\dagger$} \\
\cmidrule(lr){2-4} \cmidrule(lr){5-7}
  & Total & Veri & Non
  & Total & Veri & Non \\
\midrule
Train & 3,324 & 64\% & 36\% & 1,757 & 59\% & 41\% \\
Dev   & 307 & 64\% & 36\% & 196 & 59\% & 41\% \\
Test  & 251  & 59\% & 41\% & 816 & 64\% & 36\% \\
\bottomrule
\end{tabular}
\par
{\raggedright\scriptsize $^\dagger$ PoliClaim has no official dev split; dev set is derived from training data via stratified random split (90/10, seed=42).\par}
\end{table}

Our experiments use two complementary datasets with binary verifiability labels. Table~\ref{tab:dataset_stats} summarizes the label distributions; both datasets show moderate class imbalance, with verifiable claims at roughly 59--64\%.

\textbf{PoliClaim} contains sentence-level extracts from political speeches and debates, covering topics such as infrastructure, public policy, and economic issues. The texts are formal and grammatically standardized, with an average length of 22 words. Since each instance is extracted at the sentence level, surrounding discourse is often missing, which can make entity disambiguation difficult.

\textbf{CT22} consists of tweet-level texts from CLEF-2022 CheckThat! Lab Task~1B, focusing on COVID-19-related topics. We use its binary verifiability labels rather than check-worthiness annotations. Tweets average 36 words and are more informal. Unlike PoliClaim, CT22 tweets often form self-contained narratives, although they may mix personal opinions, anecdotes, and factual claims within the same post.

\subsection{Models}\label{sec:model}
We evaluate two model types, encoder-based and decoder-based architectures, under three learning settings. \textbf{Fine-tuning} updates model parameters on task-specific training data. \textbf{In-context learning (ICL)} evaluates models without parameter updates and includes \textbf{zero-shot learning}, where no examples are provided, and \textbf{few-shot learning}, where three examples are included in the prompt. This design allows us to assess both adaptability through parameter updates and task generalization without retraining. For encoder-only models, we use BERT-base~\citep{DBLP:journals/corr/abs-1810-04805} and RoBERTa-large~\citep{DBLP:journals/corr/abs-1907-11692}. For decoder-only models, we use two recent open-source LLMs, Llama-3-8B-Instruct (Llama3)~\citep{llama3modelcard} and Mistral-7B-Instruct-v0.2 (Mistral)~\citep{jiang2023mistral7b}, selected for their strong performance on language tasks. For in-context learning settings, we also include GPT-4o~\citep{gpt4o2025}, which performs well in such scenarios. This setup enables a direct comparison of model architectures for the verifiable claim detection task.

\subsubsection{Baseline Models}\label{sec:baseline-model}
Models that take only the claim as input serve as our baselines. For BERT and RoBERTa, we use standard tokenization with \texttt{[CLS]} and \texttt{[SEP]} tokens, and we use the \texttt{[CLS]} representation for classification. Inputs are padded or truncated to 128 tokens. We add a learnable attention layer on top of the final hidden states so that the model can focus on key parts of the claim. These models are fine-tuned for verifiability detection. For Llama3 and Mistral, we use the default tokenization from their Hugging Face implementations. Inputs follow a standard instruction prompt\footnote{All prompts are included in the supplementary material available in the repository.}, and we do not manually insert special tokens, since the models handle formatting internally. For in-context learning experiments, Llama3 and Mistral are deployed locally and run using Ollama~\citep{ollama2025}, while GPT-4o~\citep{gpt4o2025} is accessed through its official API.

\subsubsection{ContextClaim Models}
Models that use both the claim and its additional context implement the full ContextClaim, abbreviated as CC. Depending on the context generator, we label these models as CC-G4o when using GPT-4o or CC-M when using Mistral.

For BERT and RoBERTa, the claim and context are encoded by two separate instances of the same pretrained encoder, producing independent hidden state sequences. We then apply a multi-head cross-attention layer (with eight heads) in which the \texttt{[CLS]} representation of the claim serves as the query, while the full token-level hidden states of the context serve as the keys and values. This allows the claim representation to selectively attend to the most relevant parts of the retrieved context. The resulting attention output is concatenated with the claim \texttt{[CLS]} vector and passed through a linear fusion layer with ReLU activation before the final classification head.

For Llama3, Mistral, and GPT-4o, we extend the baseline prompt to include both the claim and its context as a dual input.

\subsection{Implementation Details}
All experiments were run on an NVIDIA A100 80GB GPU with CUDA 11.8, PyTorch 2.6.0, and Transformers 4.49.0. Encoder-only models were fine-tuned with a custom training loop, while decoder-only models were trained using HuggingFace \texttt{SFTTrainer} with 4-bit quantization and LoRA-based PEFT. Encoder-only results are averaged over five random seeds, and decoder-only results over three seeds due to computational cost. For encoder-only models, hyperparameters were optimized on the CT22 development set using Optuna/TPE and then transferred to PoliClaim without dataset-specific tuning; decoder-only settings were selected empirically for efficiency and performance. The retrieval module uses fixed parameters across all experiments: \(p=5\) candidate extracts, \(\alpha=0.8\), \(\beta=0.2\), and \(\theta=0.5\), none of which were tuned on the development set. Zero-shot and few-shot settings use fixed prompts, with three class-balanced examples selected deterministically from the training set.

\paragraph{Pipeline Cost}
The retrieval pipeline processes roughly three claims per minute, involving GPT-4o summarization (one call per claim); NER is local and near-instantaneous, and Wikipedia API calls add about one second per entity. For CT22 (3,324 training claims), the full summarization pass takes approximately 18 hours and costs roughly \$5 USD. The resulting summaries are generated once and cached for reuse across all downstream models and settings.

\paragraph{Evaluation Metrics} 
We report binary F1 over the verifiable (positive) class as the primary metric, with accuracy as a complementary measure. F1 is our primary evaluation metric considering the skewed class imbalance of both datasets (verifiable: 59–64\%; Table 1), while accuracy is retained for direct comparison with the CheckThat! 2022 Task 1B leaderboard \cite{nakov2022overview} and AFaCTA \cite{ni2024afacta}, which report accuracy only.

\paragraph{Research Questions}
We organize the results (Section~\ref{sec:discuss}) and analyses (Section~\ref{sec:analysis}) around four research questions:
\textbf{RQ1}: Does retrieved context improve verifiable claim detection, and under what conditions?
(\textbf{RQ1a}) Overall effectiveness across datasets;
(\textbf{RQ1b}) the role of model architecture;
(\textbf{RQ1c}) the role of learning setting.
\textbf{RQ2}: What role does LLM-based context summarization play?
(\textbf{RQ2a}) Ablation of raw retrieval vs. summarized context; % Ablation of raw retrieval vs./ summarized context;
(\textbf{RQ2b}) human evaluation of summary quality.
\textbf{RQ3}: When does retrieved context fail, and what error trade-offs does it introduce?
\textbf{RQ4}: Can context retrieved at the detection stage transfer to downstream claim verification?

\section{Results and Discussion}\label{sec:discuss}

\begin{table}[t]
\centering
\caption{Fine-tuning results on CT22 and PoliClaim (mean $\pm$ std).
Bold: column-wise best. Underline: overall best per metric.}
\label{tab:ft_results}
\setlength{\tabcolsep}{5pt}
\footnotesize
\begin{tabular}{ll rr rr}
\toprule
& & \multicolumn{2}{c}{CT22} & \multicolumn{2}{c}{PoliClaim} \\
\cmidrule(lr){3-4}\cmidrule(lr){5-6}
Model & System & Acc & F1 & Acc & F1 \\
\midrule
\multirow{3}{*}{BERT}
  & Baseline & 68.29$_{\pm\text{1.37}}$ & 76.45$_{\pm\text{0.72}}$ & 82.77$_{\pm\text{0.41}}$ & 87.07$_{\pm\text{0.32}}$ \\
  & CC-G4o   & \textbf{70.20}$_{\pm\text{1.77}}$ & \textbf{77.18}$_{\pm\text{1.06}}$ & 84.02$_{\pm\text{0.97}}$ & 87.94$_{\pm\text{0.59}}$ \\
  & CC-M     & 69.48$_{\pm\text{0.83}}$ & 76.89$_{\pm\text{0.28}}$ & \textbf{84.09}$_{\pm\text{1.48}}$ & \textbf{88.06}$_{\pm\text{0.81}}$ \\
\midrule
\multirow{3}{*}{RoBERTa}
  & Baseline & 70.84$_{\pm\text{1.94}}$ & 78.53$_{\pm\text{0.50}}$ & 85.78$_{\pm\text{0.92}}$ & 89.42$_{\pm\text{0.67}}$ \\
  & CC-G4o   & 71.95$_{\pm\text{3.40}}$ & 78.64$_{\pm\text{1.85}}$ & \underline{\textbf{86.45}}$_{\pm\text{0.53}}$ & \underline{\textbf{89.71}}$_{\pm\text{0.20}}$ \\
  & CC-M     & \underline{\textbf{73.63}}$_{\pm\text{0.33}}$ & \underline{\textbf{79.54}}$_{\pm\text{0.65}}$ & 85.93$_{\pm\text{0.49}}$ & 89.30$_{\pm\text{0.29}}$ \\
\midrule
\multirow{3}{*}{Llama3}
  & Baseline & 62.95$_{\pm\text{1.05}}$ & 73.20$_{\pm\text{0.82}}$ & 69.12$_{\pm\text{0.44}}$ & 77.50$_{\pm\text{0.28}}$ \\
  & CC-G4o   & 62.95$_{\pm\text{0.00}}$ & 75.33$_{\pm\text{0.00}}$ & \textbf{72.51}$_{\pm\text{0.14}}$ & \textbf{80.11}$_{\pm\text{0.07}}$ \\
  & CC-M     & \textbf{63.21}$_{\pm\text{0.23}}$ & \textbf{75.72}$_{\pm\text{0.11}}$ & 72.02$_{\pm\text{0.26}}$ & 79.74$_{\pm\text{0.16}}$ \\
\midrule
\multirow{3}{*}{Mistral}
  & Baseline & 60.56$_{\pm\text{0.40}}$ & 74.37$_{\pm\text{0.21}}$ & \textbf{84.40}$_{\pm\text{0.39}}$ & \textbf{88.05}$_{\pm\text{0.33}}$ \\
  & CC-G4o   & \textbf{72.64}$_{\pm\text{1.61}}$ & \textbf{78.19}$_{\pm\text{1.02}}$ & 80.92$_{\pm\text{0.39}}$ & 85.43$_{\pm\text{0.31}}$ \\
  & CC-M     & 68.66$_{\pm\text{0.23}}$ & 74.78$_{\pm\text{0.47}}$ & 80.80$_{\pm\text{0.26}}$ & 85.15$_{\pm\text{0.25}}$ \\
\bottomrule
\end{tabular}
\par
\raggedright\scriptsize
Results for BERT and RoBERTa are averaged over 5 random seeds; Llama3 and Mistral over 3 seeds due to computational constraints.
\end{table}

\begin{table}[t]
\centering
\caption{Zero-shot and few-shot results on CT22 and PoliClaim.
Bold: column-wise best. Underline: overall best per metric.}
\label{tab:icl_results}
\setlength{\tabcolsep}{2.8pt}
\footnotesize
\begin{tabular}{ll rr rr rr rr}
\toprule
& & \multicolumn{4}{c}{CT22} & \multicolumn{4}{c}{PoliClaim} \\
\cmidrule(lr){3-6}\cmidrule(lr){7-10}
& & \multicolumn{2}{c}{Zero-shot} & \multicolumn{2}{c}{Few-shot} & \multicolumn{2}{c}{Zero-shot} & \multicolumn{2}{c}{Few-shot} \\
\cmidrule(lr){3-4}\cmidrule(lr){5-6}\cmidrule(lr){7-8}\cmidrule(lr){9-10}
Model & System & Acc & F1 & Acc & F1 & Acc & F1 & Acc & F1 \\
\midrule
\multirow{3}{*}{Llama3}
  & Baseline & 60.16 & 74.87 & \textbf{71.31} & \textbf{79.78} & 69.24 & 79.71 & 72.55 & 80.42 \\
  & CC-G4o   & 62.15 & 75.83 & 68.53 & 78.59 & 70.96 & 80.23 & 74.26 & \textbf{81.87} \\
  & CC-M     & \textbf{62.55} & \textbf{76.02} & 67.33 & 78.07 & \textbf{71.81} & \textbf{80.87} & \textbf{74.51} & 80.74 \\
\midrule
\multirow{3}{*}{Mistral}
  & Baseline & \textbf{67.73} & \textbf{78.28} & 63.75 & 76.49 & \underline{\textbf{79.66}} & 84.28 & 71.08 & 81.21 \\
  & CC-G4o   & 65.34 & 77.04 & \textbf{70.52} & \textbf{78.36} & 78.43 & 84.12 & \underline{\textbf{76.35}} & \underline{\textbf{82.75}} \\
  & CC-M     & 65.34 & 77.04 & 69.32 & 78.31 & 79.04 & \underline{\textbf{84.50}} & 74.39 & 80.45 \\
\midrule
\multirow{3}{*}{GPT-4o}
  & Baseline & 72.51 & 79.88 & \underline{\textbf{79.28}} & \underline{\textbf{81.94}} & 76.23 & 77.60 & \textbf{67.28} & \textbf{65.55} \\
  & CC-G4o   & 74.10 & 80.94 & \underline{\textbf{79.28}} & 81.82 & \textbf{77.82} & \textbf{79.46} & 65.20 & 62.53 \\
  & CC-M     & \underline{\textbf{74.50}} & \underline{\textbf{81.18}} & 77.69 & 80.69 & 77.82 & 79.64 & 65.07 & 62.45 \\
\bottomrule
\end{tabular}
\par
\raggedright\scriptsize
No variance reported; Results are deterministic given fixed prompts and greedy decoding.
\end{table}

\begin{table}[t]
\centering
\caption{Comparison with published systems on CT22 and PoliClaim.}
\label{tab:sota_comparison}
\begin{tabular}{llc}
\toprule
\textbf{System} & \textbf{Dataset} & \textbf{Acc (\%)}  \\
\midrule
Majority Baseline       & CT22      & 59.36  \\
CheckThat!'22 Best~\citep{agresti2022polimi} & CT22 & 76.10 \\
ContextClaim CC-G4o (Ours) & CT22      & \textbf{79.28} \\
\midrule
Majority Baseline       & PoliClaim    & 63.85  \\
AFaCTA~\citep{ni2024afacta} & PoliClaim & 86.27 \\
ContextClaim CC-G4o (Ours) & PoliClaim & \textbf{86.45} \\
\bottomrule
\end{tabular}
\par
{\raggedright\scriptsize Accuracy only, as reported in the original publication.\par}
\end{table}

\subsection{Overall Performance (RQ1a)}\label{sec:overall_performance}

Table~\ref{tab:ft_results} reports parameter-updated fine-tuning experiments, whereas Table~\ref{tab:icl_results} evaluates frozen decoder models under zero-shot and few-shot prompting. For fine-tuned models, each baseline is compared against two ContextClaim variants; for in-context learning, GPT-4o is additionally included. Overall, incorporating retrieved context frequently improves performance, but the magnitude and consistency of these gains vary substantially across models and settings.

To quantify these trends, we evaluate improvements separately for Accuracy and F1 at the level of configuration-metric pairs (10 configurations $\times$ 2 metrics per dataset). Under this protocol, CC-G4o outperforms the corresponding baseline in 13/20 cases on Accuracy and 14/20 cases on F1, with improvements occurring somewhat more consistently on CT22 than on PoliClaim. In contrast, CC-M exhibits less stable behavior, producing both gains and degradations across configurations. These aggregate results indicate that context augmentation can be beneficial, but its effectiveness is not uniform across all settings.

At the level of individual configurations, the strongest fine-tuning result is achieved by RoBERTa with CC-G4o on PoliClaim (89.71\% F1), while we observe the largest improvement on CT22 for Mistral, whose accuracy increases by over 12 percentage points with CC-G4o. As shown in Table~\ref{tab:sota_comparison}, these configurations are competitive with prior work, with CC-G4o outperforming the CheckThat! 2022 winning system on CT22 and slightly surpassing AFaCTA on PoliClaim. These comparisons follow the standard evaluation protocol of prior work, which reports performance in terms of accuracy.

A recurring pattern across both tables is that models with weaker baselines tend to benefit more from added context, whereas stronger models exhibit smaller or less consistent gains. At the same time, context augmentation can occasionally be detrimental when the injected information is misaligned with the task or with the model's internal decision process. For example, fine-tuned Mistral on PoliClaim shows a noticeable drop in F1 under context augmentation, suggesting that additional information does not necessarily improve performance in all cases.

\begin{figure}[t]
    \centering
    \includegraphics[width=0.9\linewidth]{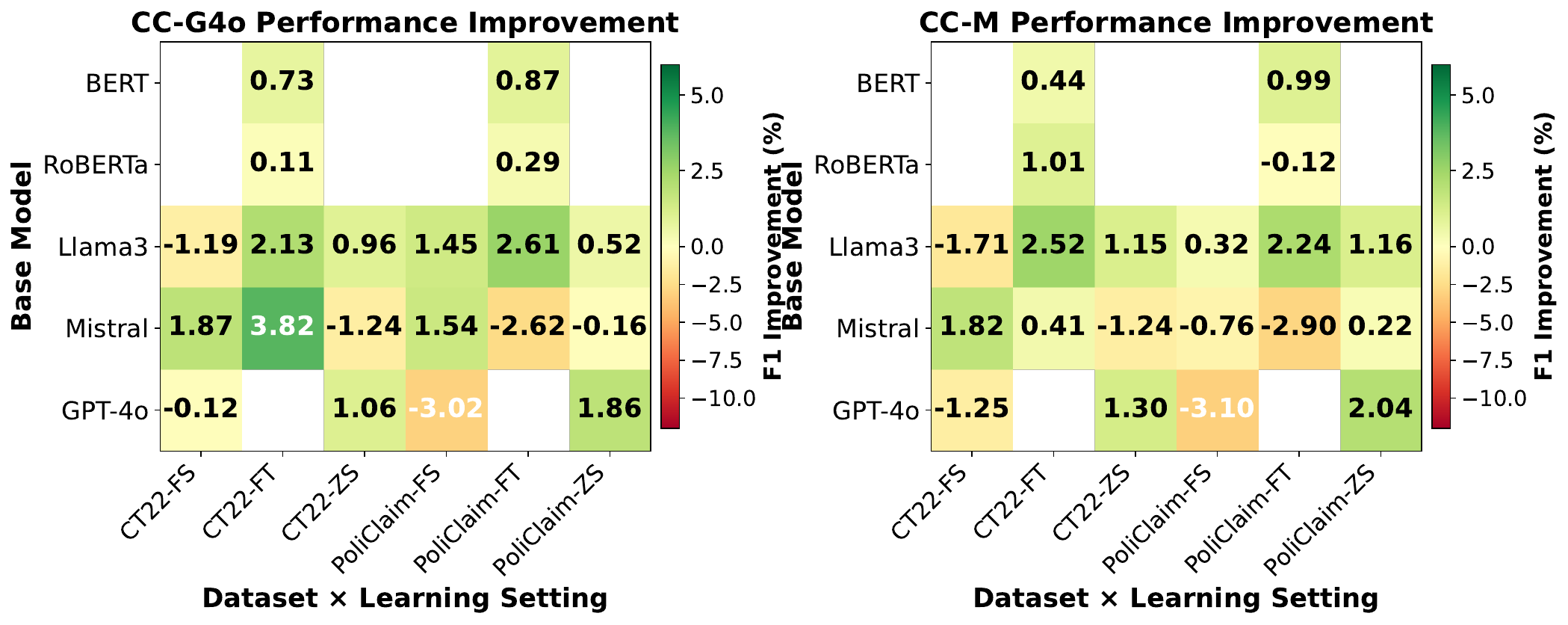}
    \caption{Performance Improvement on two Datasets with three Settings.}
    \label{fig:heatmap}
\end{figure}

\begin{figure}[t]
    \centering
    \includegraphics[width=0.9\linewidth]{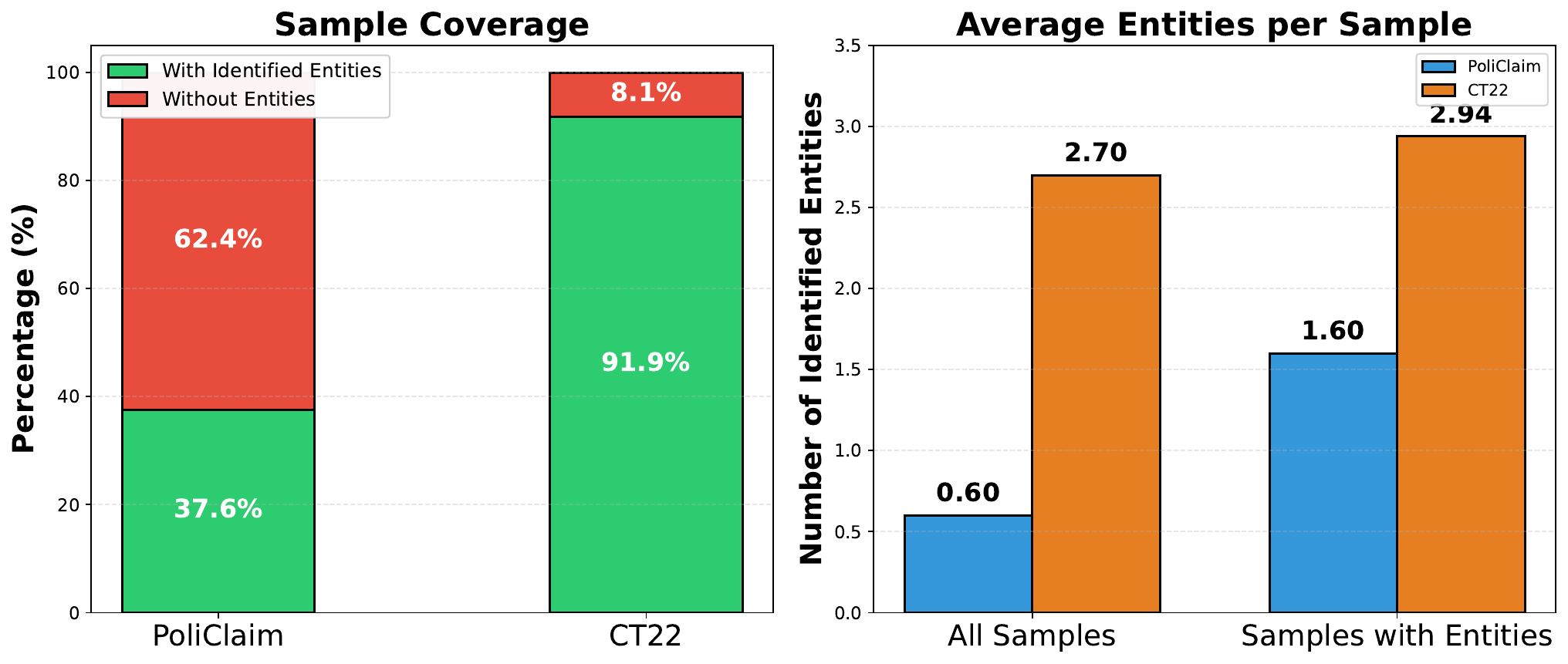}
    \caption{Entity Coverage and Quantity Comparison of PoliClaim and CT22.}
    \label{fig:coverage_and_quantity}
\end{figure}

\begin{figure}[t]
    \centering
    \includegraphics[width=0.9\linewidth]{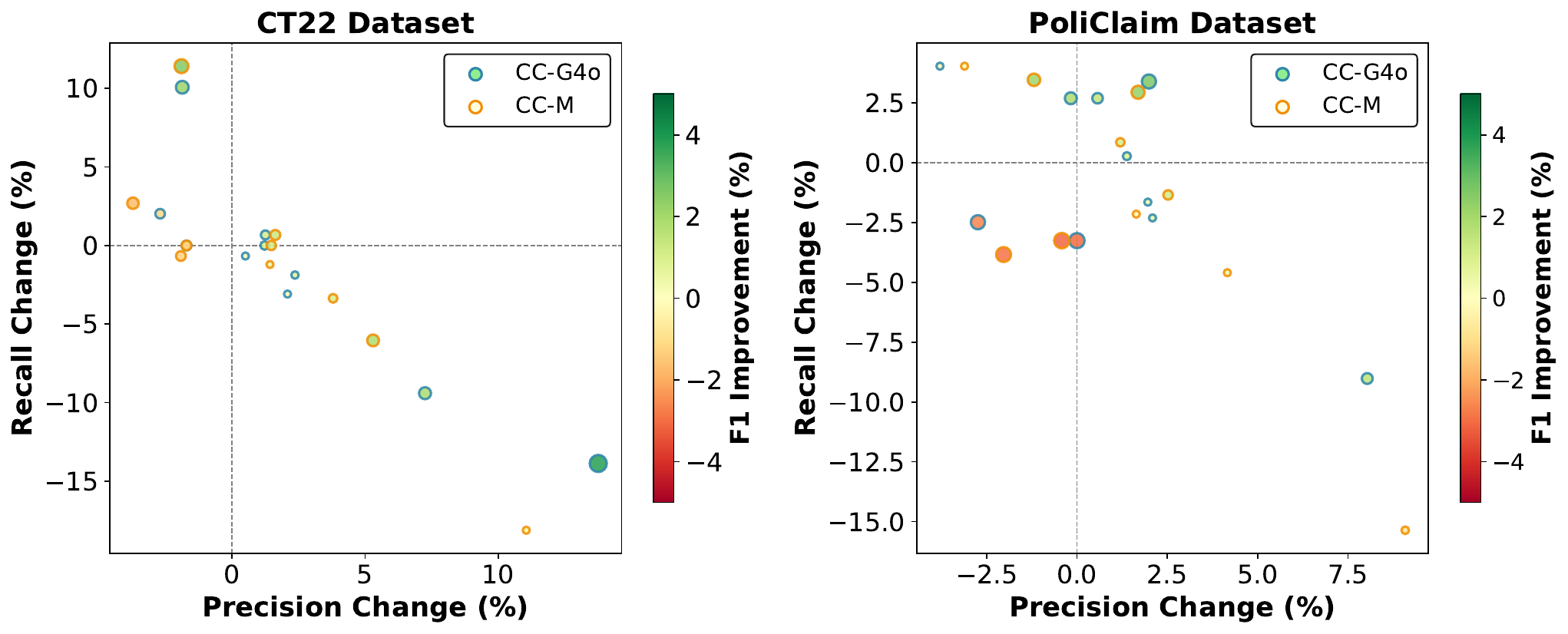}
    \caption{Precision-Recall Trade-off on Two Datasets.}
    \label{fig:precision_recall}
\end{figure}

Finally, context augmentation is more consistent on CT22 than on PoliClaim, which shows a more heterogeneous mix of gains and losses (Figure~\ref{fig:heatmap}). This pattern is consistent with the substantially higher entity coverage on CT22 (91.9\% vs. 37.6\%; Figure~\ref{fig:coverage_and_quantity}), which provides more opportunities for retrieval-based augmentation, although coverage alone does not capture the correctness of entity identification or retrieval. The precision-recall analysis in Figure~\ref{fig:precision_recall} shows a similar contrast, with CT22 configurations mostly yielding positive or near-zero F1 changes despite varied trade-offs, while PoliClaim exhibits smaller but more mixed shifts, consistent with its lower entity coverage and greater dependence on missing discourse context.

\subsection{Effects of Model Architectures (RQ1b)}

\begin{figure}[t]
    \centering
    \includegraphics[width=0.9\linewidth]{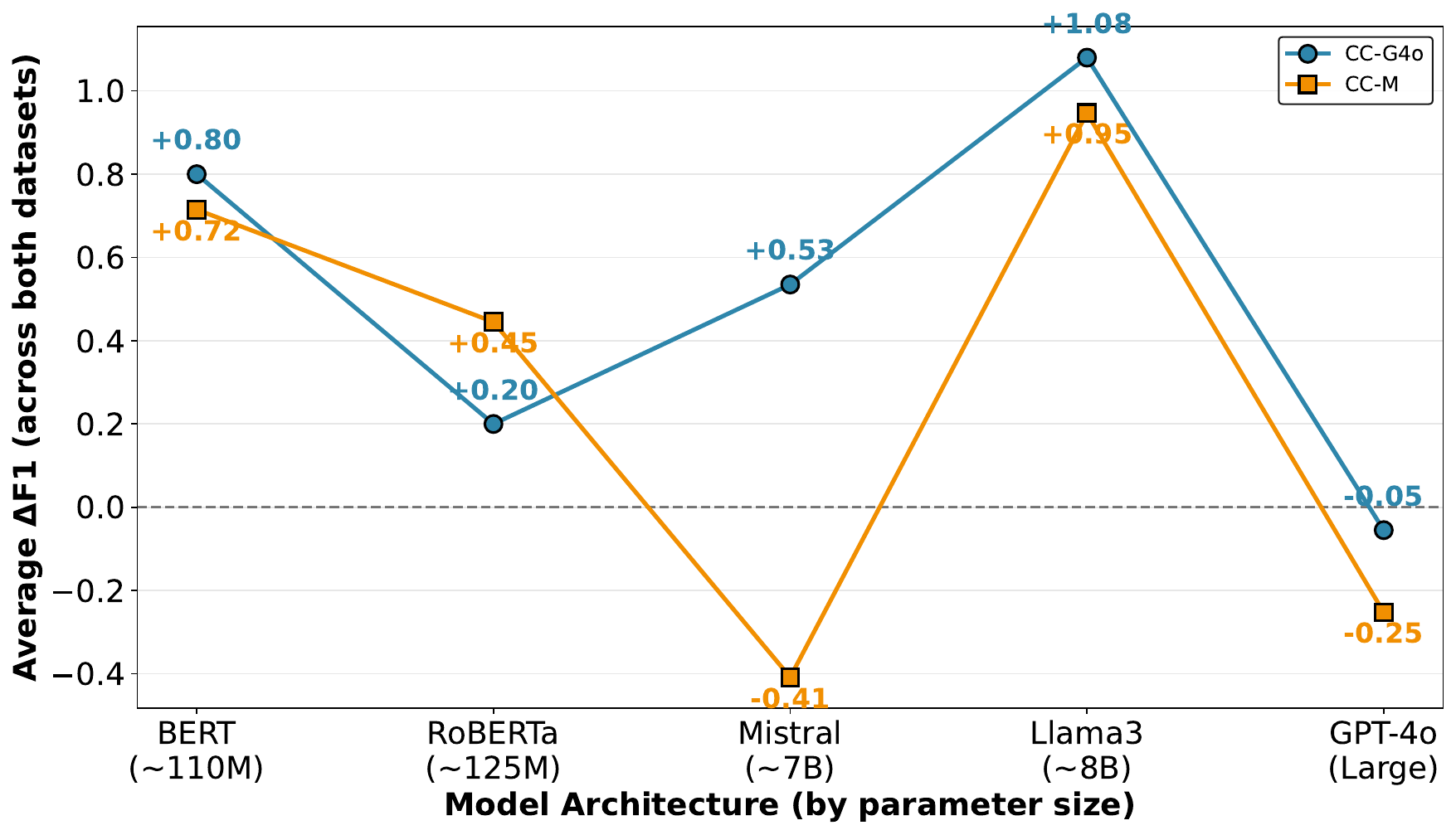}
    \caption{Model Architecture Effects on Contextual Information Benefits.}
    \label{fig:model_architecture}
\end{figure}

Figure~\ref{fig:model_architecture} plots the average difference in performance (measured as $\Delta$F1) from context augmentation across model architectures. Rather than revealing a clear relationship with model scale, the results point to a more fundamental factor on how stably different models integrate externally provided context. For most models, including BERT, Llama3, and GPT-4o, the two context variants produce broadly similar effects, suggesting that these models can incorporate additional information without substantial disruption. In these cases, the specific choice of context generator has only a limited impact on performance. Mistral, however, exhibits a qualitatively different behavior. While CC-G4o yields moderate gains, CC-M leads to a substantial performance drop, indicating that the injected context can actively interfere with the model's predictions. This points to a failure mode of context augmentation in which unstable integration of external information can cause additional context to degrade performance rather than improve it.

Taken together, these findings indicate that the effectiveness of context augmentation is governed less by model scale alone than by the stability of context integration. Models that can robustly incorporate external information tend to benefit more consistently, while those that are more sensitive to misalignment or noise exhibit highly variable outcomes.

\subsection{Effects of Learning Settings (RQ1c)}

\begin{figure}[t]
    \centering
    \includegraphics[width=0.9\linewidth]{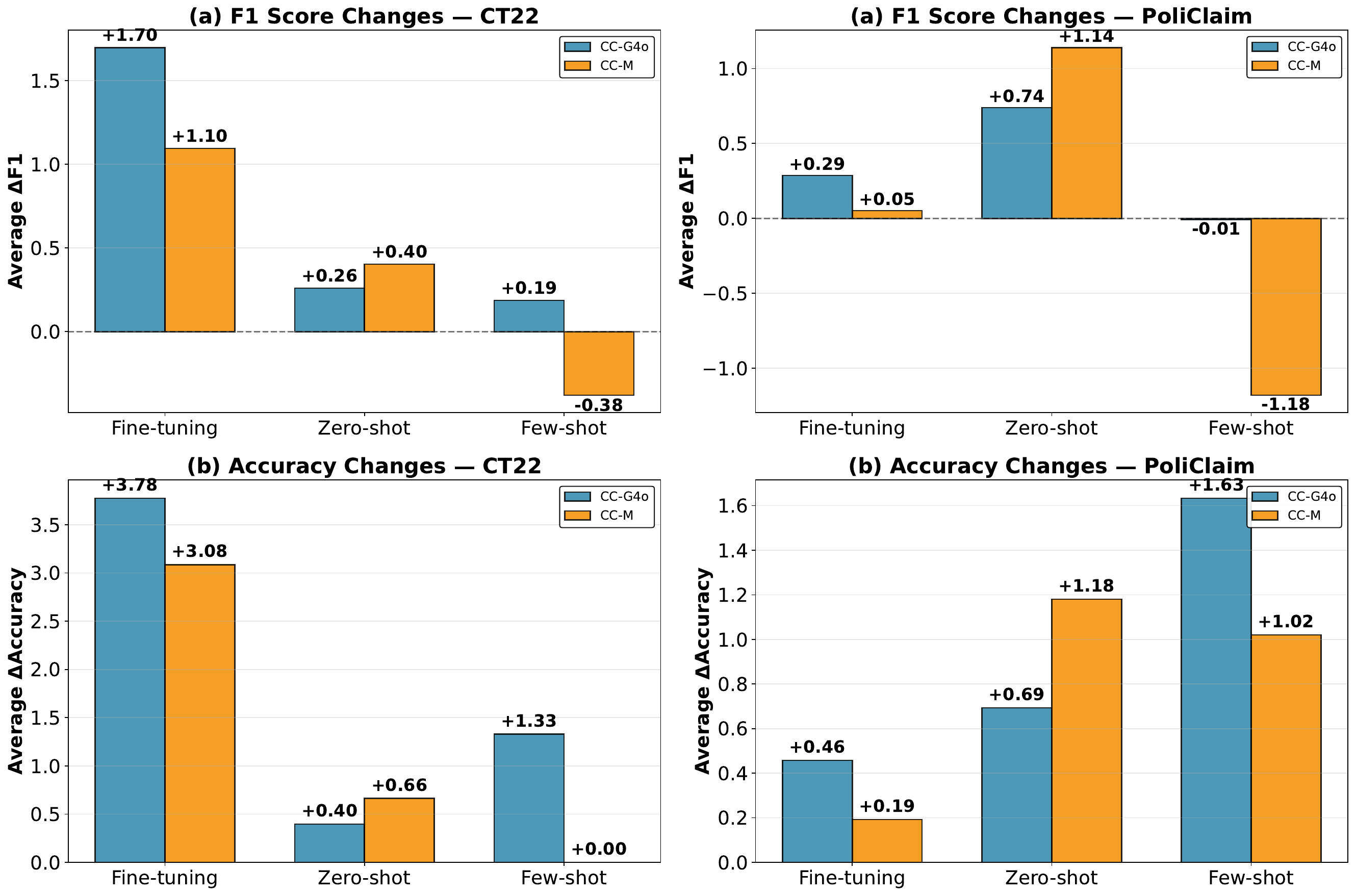}
    \caption{Learning Setting Comparison.}
    \label{fig:learning_settings}
\end{figure}

ContextClaim's effectiveness also varies considerably across learning settings, with fine-tuning and zero-shot configurations benefiting more consistently from context augmentation, while few-shot learning presents a more complicated picture. Figure~\ref{fig:learning_settings} summarises these patterns across both datasets.

Fine-tuning tends to provide the most stable environment for context integration. On CT22, both CC-G4o and CC-M deliver consistent F1 gains across all four model configurations, with average improvements of roughly 1 to 2 F1 points and over three accuracy points. On PoliClaim, the gains are more modest, reflecting the additional challenge posed by the shorter, more fragmentary nature of political speech excerpts. Overall, fine-tuning provides a more favorable setting for benefiting from external context than prompt-based settings alone.

Zero-shot learning tells a somewhat different story, particularly for PoliClaim, where the benefit of context exceeds that observed under fine-tuning. Both CC-G4o and CC-M improve F1 on PoliClaim by roughly one point, and the gains on CT22, though smaller, remain generally positive. One possible explanation is that, in the absence of demonstrations, retrieved context serves as a form of decision anchor, allowing models that would otherwise rely on surface-level cues to use the external summary as grounding for more calibrated verifiability judgments.

Few-shot learning is where context augmentation shows its limitations most clearly. On CT22, the F1 changes from context are near zero or slightly negative, contrasting sharply with the consistent gains under fine-tuning. On PoliClaim, accuracy improves for both variants, yet F1 declines, suggesting a shift in the precision-recall balance rather than a straightforward degradation. When few-shot examples already provide a latent classification cue, extra context may act as a competing signal rather than a complementary one. Taken together, these results suggest that the impact of context augmentation depends not only on the information provided, but also on whether the learning setting allows that information to stabilize or disrupt the model's decision process.

\section{Analysis}\label{sec:analysis}
\subsection{Summarization Ablation (RQ2a)}

We assess the role of LLM-based context summarization by comparing the baseline (BL), ContextClaim with raw retrieved extracts (CC-Raw), and ContextClaim with LLM-summarized context (CC-G4o and CC-M) across Llama3 and Mistral under all three learning settings. Table~\ref{tab:ablation} reports the incremental gains at each transition.

The effect of adding raw retrieved context is highly inconsistent across models, datasets, and learning settings. As shown in Table~\ref{tab:ablation}, individual configurations range from large gains (e.g., Mistral fine-tuned on CT22 gains over 11 accuracy points from raw context) to substantial drops (e.g., Llama3 few-shot on CT22 loses over 7 accuracy points). On PoliClaim, the picture is similarly mixed, with some configurations improving on one metric while declining on the other. These results suggest that raw retrieved passages are an unreliable form of augmentation, as they may provide useful evidence in some cases but introduce noise or conflicting information in others.

LLM-based summarization often mitigates this instability. The clearest case is the CT22 few-shot setting, where raw context causes the largest degradation but summarization recovers most of the loss, with both CC-G4o and CC-M producing accuracy gains exceeding the $|\Delta|>3.5$ threshold over CC-Raw (Table~\ref{tab:ablation}). Beyond these recovery cases, summarization can also provide smaller additional gains when raw context is already helpful. Overall, these patterns indicate that summarization serves primarily as a filtering step, converting noisy retrieved passages into more focused and model-usable context.

The contribution of summarization also varies by dataset. On CT22, the largest gains from the CC-Raw$\rightarrow$CC-G4o/CC-M transition are concentrated in few-shot settings, where raw context is most harmful. On PoliClaim, the benefits are more distributed across settings and are often more visible in F1 than in accuracy. For example, under zero-shot, Llama3 and Mistral both obtain positive F1 gains from summarization even though the BL$\rightarrow$CC-Raw transition is neutral or adverse on at least one metric. The two summarization models behave similarly in many cases, although CC-G4o produces slightly more large-magnitude recovery cases than CC-M, suggesting a modest advantage in the most difficult settings. This pattern holds even when Mistral serves as both summarizer and classifier (Mistral + CC-M), providing no evidence of a self-reinforcing advantage from the shared model.

Taken together, these results show that retrieval alone is not a consistently reliable basis for context augmentation. Raw extracts can be either beneficial or disruptive, whereas LLM-based summarization is the component that most often stabilizes and refines the contribution of retrieved evidence, particularly when supervision is limited. Section~\ref{sec:error_analysis} further identifies irrelevant retrieved context and missing context among false-negative cases, which is consistent with the unstable effects of raw retrieval observed here.

\begin{table}[t]
\centering
\caption{Performance Gain ($\Delta$) from ContextClaim Components.}
\label{tab:ablation}
\setlength{\tabcolsep}{3pt}
\begin{tabular}{@{}ll|rr|rr|rr@{}}
\toprule
\multirow{2}{*}{\textbf{Setting}} & \multirow{2}{*}{\textbf{Model}} & \multicolumn{2}{c|}{\textbf{BL$\rightarrow$Raw}} & \multicolumn{2}{c|}{\textbf{Raw$\rightarrow$G4o}} & \multicolumn{2}{c}{\textbf{Raw$\rightarrow$M}} \\
\cmidrule(lr){3-4} \cmidrule(lr){5-6} \cmidrule(lr){7-8}
& & \textbf{Acc} & \textbf{F1} & \textbf{Acc} & \textbf{F1} & \textbf{Acc} & \textbf{F1} \\
\midrule
\multicolumn{8}{c}{\textit{CT22}} \\
\midrule
FT & Llama3   & -0.80 & +1.73 & +0.80 & +0.40 & +1.06 & +0.79 \\
   & Mistral  & \textbf{+11.02} & \textbf{+3.61} & +1.06 & +0.21 & -2.92 & -3.20 \\
ZS & Llama3   & +3.19 & +1.42 & -1.20 & -0.46 & -0.80 & -0.27 \\
   & Mistral  & -0.80 & -0.65 & -1.59 & -0.59 & -1.59 & -0.59 \\
FS & Llama3   & \textbf{-7.17} & -3.22 & \textbf{+4.39} & +2.03 & +3.19 & +1.51 \\
   & Mistral  & +1.59 & +0.68 & \textbf{+5.18} & +1.19 & \textbf{+3.98} & +1.14 \\
\midrule
\multicolumn{8}{c}{\textit{PoliClaim}} \\
\midrule
FT & Llama3   & +0.24 & -0.01 & +3.15 & +2.62 & +2.66 & +2.25 \\
   & Mistral  & -3.35 & -2.55 & -0.13 & -0.07 & -0.25 & -0.35 \\
ZS & Llama3   & +3.43 & -1.13 & -1.71 & +1.65 & -0.86 & +2.29 \\
   & Mistral  & -2.70 & -3.16 & +1.47 & +3.00 & +2.08 & +3.38 \\
FS & Llama3   & +1.47 & -0.01 & +0.24 & +1.46 & +0.49 & +0.33 \\
   & Mistral  & +2.45 & +0.27 & +2.82 & +1.27 & +0.86 & -1.03 \\
\bottomrule
\end{tabular}
\par
{\raggedright\scriptsize \textbf{Bold}: $|\Delta| > 3.5$. FT: Fine-tuning, ZS: Zero-shot, FS: Few-shot. BL: Baseline, Raw: CC-Raw, G4o: CC-G4o, M: CC-M\par}
\end{table}

\subsection{Human Evaluation (RQ2b)}\label{sec:human_eval}

\begin{table}[t]
\centering
\caption{Human evaluation results for context summary quality. Ratings on 1--3 scale
(1=Poor, 2=Acceptable, 3=Good). No significant differences between systems (paired t-test, all $p>0.05$).}
\label{tab:human_eval}
\begin{tabular}{lcccc}
\toprule
\textbf{Dimension} & \textbf{GPT-4o} & \textbf{Mistral} & \textbf{Diff} & \textbf{p-value} \\
\midrule
Relevance         & 2.93 $\pm$ 0.03 & 2.88 $\pm$ 0.04 & +0.06 & 0.21 \\
Signal Clarity    & 1.80 $\pm$ 0.08 & 1.74 $\pm$ 0.07 & +0.06 & 0.41 \\
Usefulness        & 1.62 $\pm$ 0.07 & 1.58 $\pm$ 0.07 & +0.03 & 0.59 \\
\midrule
\textbf{Overall}  & 2.12 $\pm$ 0.90 & 2.07 $\pm$ 0.89 & +0.05 & -- \\
\bottomrule
\end{tabular}
\end{table}

We conducted a human evaluation study to assess the quality of generated context summaries beyond downstream ContextClaim performance. Three trained annotators independently evaluated 40 randomly sampled claims from the CT22 dataset, each paired with summaries generated by GPT-4o and Mistral. Summary quality was evaluated along three dimensions. Relevance captures how directly a summary refers to the entities and events mentioned in the claim. Signal clarity reflects whether the summary provides explicit cues that are useful for verifiability assessment. Usefulness measures annotators' confidence in determining claim verifiability after reading the summary. Each dimension was rated on a three-point scale, where 1 indicates Poor, 2 indicates Acceptable, and 3 indicates Good.\footnote{Annotation guidelines are included in the supplementary material available in the repository.}

Inter-annotator agreement analysis showed fair agreement across the three dimensions. Fleiss' $\kappa$ was 0.32 for GPT-4o and 0.27 for Mistral on average, which is deemed acceptable for exploratory research involving subjective evaluation \citep{landis1977measurement}. As shown in Table~\ref{tab:human_eval}, both systems received very high relevance scores, with means of 2.93 for GPT-4o and 2.88 for Mistral. This indicates that the generated summaries are generally well aligned with claim content. By contrast, scores for signal clarity and usefulness were substantially lower. GPT-4o achieved 1.80 on signal clarity and 1.62 on usefulness, while Mistral achieved 1.74 and 1.58, respectively. Overall scores were also close, at 2.12 for GPT-4o and 2.07 for Mistral, and paired t-tests did not identify significant differences between the two systems on any dimension.

These findings help clarify the pattern observed in the downstream experiments. The summaries produced by both models are usually relevant, but they do not consistently make verifiability signals explicit enough for human readers. In other words, the generated context often provides topical grounding without always providing decision-supporting evidence in a clear form. This helps explain why context augmentation can improve performance in many settings while still yielding heterogeneous outcomes across models, datasets, and learning regimes.

The absence of significant human-rated differences between GPT-4o and Mistral is also consistent with the architecture-level analysis. For most models, the downstream gap between CC-G4o and CC-M is relatively small, and the largest divergence is concentrated in a limited number of cases rather than being uniformly distributed. These results indicate that the downstream effect of summarization is not determined solely by average perceived quality, but also by how well the provided context interacts with a model's decision process. In this sense, summarization appears to contribute primarily by filtering and focusing retrieved evidence, while its ultimate utility depends on whether the resulting context stabilizes or perturbs prediction.

More broadly, the relatively low scores on signal clarity and usefulness indicate that there remains substantial room to improve how retrieved evidence is transformed into task-relevant context. The current summaries are generally relevant, but often stop short of making the information most useful for verifiability judgments explicit. This limitation is consistent with the mixed results observed in few-shot settings and on more discourse-dependent inputs such as PoliClaim, where relevant context may still be insufficient to support stable decision-making.

\subsection{Error Analysis (RQ3)}\label{sec:error_analysis}

We conduct a detailed error analysis to understand the limitations of the ContextClaim paradigm, focusing on two representative model configurations, RoBERTa with CC-M (encoder-only) and Llama3 with CC-G4o (decoder-only). Comparing these two architectures allows us to distinguish paradigm-level failure modes from architecture-specific ones.

\subsubsection{Error distribution and false positive bias}

Table~\ref{tab:error_distribution} reports the error distribution for both models and their baselines on CT22. A systematic false positive bias is present across all configurations, suggesting that this is a paradigm-level issue rather than an artefact of a single model. However, the severity and direction of this bias differ markedly by architecture. For Llama3, adding CC-G4o context dramatically amplifies the bias. The FP/FN ratio rises from 3.09 to 12.29, with FP increasing from 71 to 86 while FN drops from 23 to 7. For RoBERTa, the effect is reversed. CC-M suppresses the bias, with the FP/FN ratio falling from 4.85 to 2.67, as FP decreases from 63 to 48. This architectural divergence suggests that encoder-only models can leverage cross-attention to discriminate between entity presence and verifiability, whereas decoder-only models tend to treat contextual signals as evidence that a claim is verifiable.

\begin{table}[t]
\centering
\caption{Error Distribution by Model on CT22 (n=251).}
\label{tab:error_distribution}
\setlength{\tabcolsep}{4pt}
\small
\begin{tabular}{lcccc}
\toprule
\textbf{Model} & \textbf{FP} & \textbf{FN} & \textbf{FP/FN} & \textbf{Errors} \\
\midrule
Llama3 Baseline   & 71 (28.3\%) & 23 (9.2\%) & 3.09 & 94 (37.5\%) \\
Llama3 + CC-G4o   & 86 (34.3\%) &  7 (2.8\%) & 12.29 & 93 (37.1\%) \\
\midrule
RoBERTa Baseline  & 63 (25.1\%) & 13 (5.2\%) & 4.85 & 76 (30.3\%) \\
RoBERTa + CC-M    & 48 (19.1\%) & 18 (7.2\%) & 2.67 & 66 (26.3\%) \\
\bottomrule
\end{tabular}
\end{table}

\subsubsection{Error taxonomy}
To move beyond aggregate counts, we manually categorized all FP and FN instances across both CC models into a lightweight taxonomy. Among false positives, three types account for the majority of classifiable errors. \textit{Opinion or call-to-action} errors (27\% of FP cases) arise when a claim expresses a normative judgement or policy stance that is not fact-checkable, yet references real entities whose Wikipedia context triggers a verifiable prediction. For example, ``The bold move by POTUS to impose vaccine mandates will save countless lives.'' \textit{Humour and personal anecdote} errors (11\% of FP cases) occur when a tweet adopts a narrative surface form that superficially resembles a factual report but is not checkable, such as ``My cousin got the vaccine and three days later stepped on a lego.'' \textit{Factual-context-misleads} errors (11\% of FP cases) occur when retrieved context is accurate and entity-relevant but provides no signal about whether the claim itself is verifiable. For example, the claim ``Alaska made history this year'' is paired with geographic background about Alaska, which confirms the entity but says nothing about the vague assertion, leading the model to predict verifiable.

Among false negatives, errors are more evenly distributed across four types. \textit{Irrelevant context retrieved} (18\% of FN cases) occurs when entity extraction succeeds but the Wikipedia extract addresses a different aspect of the entity than the one the claim concerns. \textit{No context, verifiable claim missed} (18\% of FN cases) occurs when entity extraction fails entirely, leaving the model without any augmented signal. \textit{Anecdote or quote is checkable} (23\% of FN cases) covers cases where a conversational or attributed framing conceals a concrete, verifiable assertion, for instance a direct quote from a named public figure about a documented event. \textit{Metaphor masks verifiability} (14\% of FN cases) covers cases where analogical phrasing obscures a checkable factual claim, such that neither the model nor the retrieval pipeline identifies a verifiable core.

\subsubsection{Error transition analysis}
To quantify what context adds and what it costs, we align baseline and CC predictions sample by sample and classify each into four transitions. For Llama3, CC-G4o improves roughly as many samples as it regresses, but the nature of the changes differs: most improvements come from fixing false negatives where context provides verifiability evidence, while most regressions are new false positives where factual context appears to mislead the model. For RoBERTa, CC-M improves about twice as many samples as it regresses, predominantly by correcting false positives where the baseline incorrectly labelled opinions or anecdotes as verifiable. The net effect consolidates the architectural contrast: Llama3 shifts toward higher recall at the cost of precision, while RoBERTa shifts toward higher precision at a modest recall cost.

Overall, these findings show that the benefits of ContextClaim come with different error trade-offs across architectures. For decoder-only models, the primary risk is that retrieved context, even when relevant, can inflate false positives if it does not clearly indicate whether the claim is verifiable. For encoder-only models, context reduces false positives but can introduce new false negatives when retrieval is irrelevant. Both failure modes ultimately trace back to the gap identified in Section~\ref{sec:human_eval}. Generated summaries score highly on relevance but more weakly on signal clarity, providing factual background without explicitly indicating whether a claim is checkable.

\subsection{Downstream Verification (RQ4)}
\label{sec:downstream-verification}

We further ask whether the retrieved context can support downstream claim verification beyond detection. Since CT22 and PoliClaim lack veracity labels, we use FEVER~\cite{thorne2018FactExtraction}, a verification benchmark built on Wikipedia, the same knowledge source used by ContextClaim. FEVER's ``Not Enough Info'' label denotes insufficient evidence for a factual claim, which differs from the non-verifiable category in CT22 and PoliClaim; we therefore exclude these ${\sim}$3,333 instances from the 9,999-claim test set and evaluate on the remaining 6,666 SUPPORTS/REFUTES claims.

GPT-4o serves as a fixed verification classifier under two conditions: \textit{claim-only} (claim text only) and \textit{claim + context} (claim plus CC-G4o context summary). The retrieval pipeline follows the same three-stage design as in the main experiments (NER keyword extraction, sentence-transformer entity linking, and GPT-4o summarization) but retrieves from the FEVER-provided Wikipedia snapshot to align with the annotated knowledge base. GPT-4o is run with temperature 0. The goal is to isolate the marginal utility of ContextClaim context, not to compete with specialised FEVER systems.

As shown in Table~\ref{tab:downstream-fever}, adding ContextClaim context improves F1 by over one point, with the gain driven primarily by a substantial increase in recall at the cost of lower precision. This precision-recall trade-off is consistent with the false-positive tendency of decoder-only models noted in Section~\ref{sec:error_analysis}. We observe the same trend in a preliminary 1,000-instance subset, confirming robustness across sample sizes.

This experiment tests verification rather than verifiability detection, using FEVER instead of CT22 or PoliClaim, with the same retrieval algorithm differing only in Wikipedia version (FEVER's 2017 snapshot vs. the live API). The consistent gains suggest that ContextClaim context carries useful signal beyond detection, supporting retrieval reuse across fact-checking stages.

% vs.\ the live API

\begin{table}[t]
\centering
\caption{Downstream verification on FEVER (SUPPORTS vs.\ REFUTES, $n{=}6{,}666$). GPT-4o classifier is held fixed; only the input varies. F1: positive=SUPPORTS.}
\label{tab:downstream-fever}
\begin{tabular}{lccc}
\toprule
\textbf{Metric} & \textbf{Claim Only} & \textbf{Claim + CC-G4o} & $\boldsymbol{\Delta}$ \\
\midrule
Accuracy  & 88.70 & 89.34 & +0.64 \\
Precision & 91.00 & 87.97 & $-$3.03 \\
Recall    & 85.89 & 91.14 & +5.25 \\
F1         & 88.37 & 89.53 & +1.16 \\
Macro-F1   & 88.69 & 89.34 & +0.65 \\
\bottomrule
\end{tabular}
\end{table}

\section{Conclusion}
We have presented Context-Driven Claim Detection (ContextClaim), a paradigm that integrates context retrieval from a structured knowledge source into the detection stage of automated fact-checking pipelines. ContextClaim extracts entity mentions from the input claim, retrieves relevant background information from Wikipedia, and uses large language models to produce concise contextual summaries for downstream classification. We introduce two variants that differ in the summarization model used, CC-G4o (GPT-4o) and CC-M (Mistral).

Our experiments on two complementary datasets, CT22 (COVID-19 tweets) and PoliClaim (political speeches), show that context augmentation can improve verifiable claim detection, though its effectiveness varies across datasets, model architectures, and learning settings. Across the 20 model--dataset configurations we evaluate, ContextClaim variants improve over the corresponding baselines in 13 cases for accuracy and 14 for F1. The effect also varies across model architectures, with the stability of context integration appearing to matter more than model scale alone. The gains are most consistent under fine-tuning, where parameter updates allow models to learn how to integrate the additional context. In zero-shot settings, retrieved context appears to serve as a useful decision anchor for some models, while in few-shot settings the results are more mixed, suggesting that additional context may interact unpredictably with the implicit decision boundaries established by in-context demonstrations.

Our human evaluation further reveals that both summarization variants produce highly relevant context, but that relevance alone does not guarantee downstream improvement. Signal clarity and usefulness are notably lower, which may help explain why gains are not uniform. These findings suggest that the effectiveness of context augmentation depends not only on whether the retrieved information is topically related to the claim, but also on whether it provides sufficiently clear signals to support stable verifiability judgments.

Beyond the detection task, we also show that ContextClaim’s retrieved context also improves downstream claim verification on FEVER despite differences in task formulation and dataset. This suggests that early-stage retrieval can provide reusable context for both detection and downstream verification.

\paragraph{Limitations}
% \subsection*{Limitations}
Our evaluation covers two specific domains, and the effectiveness of context augmentation in domains with different entity distributions or linguistic characteristics remains to be investigated. ContextClaim relies on Wikipedia as its sole knowledge source, and coverage gaps for emerging events or less prominent entities may limit its applicability. On the summarization side, our human evaluation indicates that signal clarity remains a bottleneck; developing more targeted summarization strategies that emphasize decision-relevant information could improve reliability. Finally, because CT22 and PoliClaim lack veracity labels, our downstream verification is conducted on FEVER, and a unified dataset supporting both detection and verification would enable tighter end-to-end evaluation.

\section*{Acknowledgments}
Rrubaa Panchendrarajan is funded by the European Union and UK Research and Innovation under Grant No. 101073351 as part of Marie Skłodowska-Curie Actions (MSCA Hybrid Intelligence to monitor, promote, and analyze transformations in good democracy practices). Yufeng Li is funded by the China Scholarship Council (CSC).

\section*{GenAI Usage Disclosure}

Generative AI tools were used only to assist with minor language polishing and boilerplate code. All the text is originally written by the authors, and ultimately reviewed and approved by the authors.

\bibliographystyle{IEEEtran}
\bibliography{references}

% \vfill


\end{document}